\documentclass[11pt]{article}
\usepackage{fullpage}
\usepackage{amsmath,amsfonts,amsthm,amssymb}
\usepackage[colorlinks=true, linkcolor=red, urlcolor=blue, citecolor=gray]{hyperref}
\usepackage{url}
\usepackage{algorithm}
\usepackage{algorithmic}
\usepackage{bbm}
\usepackage{float}
\usepackage{framed}
\usepackage{enumerate}
\usepackage{color}
\usepackage[usenames,dvipsname s,svgnames,table]{xcolor}
\usepackage{array}
\usepackage[para]{footmisc}
\usepackage{epsfig}
\usepackage{framed}
\usepackage[framemethod=tikz]{mdframed}
\usepackage{wrapfig}
\makeatletter
\newcommand{\thickhline}{%
    \noalign {\ifnum 0=`}\fi \hrule height 1pt
    \futurelet \reserved@a \@xhline
}
\newcolumntype{"}{@{\hskip\tabcolsep\vrule width 1pt \hskip\tabcolsep}}
\makeatother
\newcolumntype{'}{@{\hskip\tabcolsep\vrule width 1pt}}
\makeatother
\newcolumntype{`}{@{\vrule width 1pt \hskip\tabcolsep}}
\makeatother

\DeclareMathOperator*{\E}{\mathrm{E}}
\let\Pr\relax
\DeclareMathOperator*{\Pr}{\mathrm{Pr}}

\newcommand{\whp}{w.h.p. }

\DeclareMathOperator{\Bias}{b}
\DeclareMathOperator{\Net}{\mathcal{N}}
\DeclareMathOperator{\dec}{dec}
\DeclareMathOperator{\bin}{bin}
\DeclareMathOperator{\pot}{pot}

\DeclareMathOperator{\ham}{d_H}

\newcommand{\eqdef}{\mathbin{\stackrel{\rm def}{=}}}
\newcommand{\norm}[1]{\|#1\|}

\newcommand{\bs}[1]{\boldsymbol{#1}}
\newcommand{\bv}[1]{\mathbf{#1}}

\newcommand{\set}[1]{\textcolor{black}{#1}}

\makeatletter

\newtheorem*{rep@theorem}{\rep@title}
\newcommand{\newreptheorem}[2]{%
\newenvironment{rep#1}[1]{%
 \def\rep@title{#2 \ref{##1}}%
 \begin{rep@theorem}}%
 {\end{rep@theorem}}}
\makeatother
\newtheorem{theorem}{Theorem}

\newtheorem{corollary}[theorem]{Corollary}
\theoremstyle{plain}
\newtheorem{observation}[theorem]{Observation}
\newtheorem{lemma}[theorem]{Lemma}
\newtheorem*{lemma*}{Lemma}
\newreptheorem{lemma}{Lemma}
\theoremstyle{plain}
\newtheorem{fact}[theorem]{Fact}

\newtheorem{definition}[theorem]{Definition}
\def\blackslug{\hbox{\hskip 1pt \vrule width 4pt height 8pt
    depth 1.5pt \hskip 1pt}}
\def\QED{\quad\blackslug\lower 8.5pt\null\par}

\renewcommand{\paragraph}[1]{\vspace{0.15cm}\noindent {\bf #1}}
  \usepackage{nth}
  \usepackage{intcalc}

\title{Neuro-RAM Unit with Applications to Similarity Testing and Compression in Spiking Neural Networks}

\author{
Nancy Lynch \\
   MIT \\
   \texttt{lynch@csail.mit.edu}
\and
 Cameron Musco \\
   MIT \\
  \texttt{cnmusco@mit.edu}
\and
Merav Parter\\
	 MIT \\
	\texttt{parter@mit.edu}
}

\begin{document}
\maketitle

\begin{abstract}
We study distributed algorithms implemented in a simplified biologically inspired model for \emph{stochastic spiking neural networks}. We focus on tradeoffs between computation time and network complexity, along with the role of noise and randomness in efficient neural computation.

It is widely accepted that neural spike responses, and neural computation in general, is inherently stochastic. In recent work, we explored how this stochasticity could be leveraged to solve the `winner-take-all' leader election task. 
Here, we focus on using randomness in neural algorithms for similarity testing and compression. In the most basic setting, given two $n$-length patterns of firing neurons, we wish to distinguish if the patterns are equal or $\epsilon$-far from equal.

Randomization allows us to solve this task with a very compact network, using $O \left (\frac{\sqrt{n}\log n}{\epsilon}\right)$ auxiliary neurons, which is sublinear in the input size.
 At the heart of our solution is the design of a $t$-round neural random access memory, or indexing network, which we call a  \emph{neuro-RAM}. This module can be implemented with $O(n/t)$ auxiliary neurons and is useful in many applications beyond similarity testing -- e.g. we discuss its application to compression via random projection.

Using a VC dimension-based argument, we show that the tradeoff between runtime and network size in our neuro-RAM is nearly optimal.
To the best of our knowledge, we are the first to apply these techniques to stochastic spiking networks. Our result has several implications -- since our neuro-RAM can be implemented with deterministic threshold gates, it shows that, in contrast to similarity testing, randomness does not provide significant computational advantages for this problem. It also establishes a separation between feedforward networks whose gates spike with sigmoidal probability functions, and well-studied deterministic sigmoidal networks, whose gates output real number sigmoidal values, and which can implement a neuro-RAM much more efficiently.
\end{abstract}

\thispagestyle{empty}
\clearpage
\setcounter{page}{1}

\section{Introduction}\label{sec:intro}
Biological neural networks are arguably the most fascinating distributed computing systems in our world. However, while studied extensively in the fields of computational neuroscience and artificial intelligence, they have received little attention from a distributed computing perspective.
Our goal is to study biological neural networks through the lens of distributed computing theory. We focus on
understanding tradeoffs between computation time, network complexity, and the use of randomness in implementing basic algorithmic primitives, which can serve as building blocks for high level pattern recognition, learning, and processing tasks.

\paragraph{Spiking Neural Network (SNN) Model}
We work with biologically inspired \emph{spiking neural networks} (SNNs) \cite{maass1996computational,maass1997networks,gerstner2002spiking,izhikevich2004model}, in which neurons fire in discrete pulses in synchronous rounds, in response to a sufficiently high membrane potential. This potential is induced by spikes from neighboring neurons, which can have either an excitatory or inhibitory effect (increasing or decreasing the potential). As observed in biological networks, neurons are either strictly inhibitory (all outgoing edge weights are negative) or excitatory. As we will see, this restriction can significantly affect the power of these networks. 

A key feature of our model is stochasticity -- each neuron is a probabilistic threshold unit, spiking with probability given by applying a sigmoid function to its potential. 
%
While a rich literature focuses on deterministic circuits \cite{minsky1969perceptrons,hopfield1986computing}
we employ a stochastic model as 
it is widely accepted that neural computation is stochastic \cite{allen1994evaluation,shadlen1994noise,faisal2008noise}.

\paragraph{Computational Problems in SNNs}
We consider an $n$-bit binary input vector $X$, which represents the firing status of a set of input neurons. Given a (possibly multi-valued) function $f: \{0,1\}^n \to \{0,1\}^m$, we seek to design a network of spiking neurons that converges to an output vector $Z = f(X)$  (or any $Z \in f(X)$ if $f$ is multi-valued) as quickly as possible using few auxiliary (non-input or output) neurons.

The number of auxiliary neurons used corresponds to the ``node complexity" of the network \cite{horne1994node}. Designing circuits with small node complexity has received a lot of attention -- e.g., the work of \cite{furst1984parity} on PARITY and \cite{allender1989note} on $AC_0$. Much less is known, however, on what is achievable in spiking neural networks. For most of the problems we study, there is a trivial solution that uses $\Theta(n)$ auxiliary neurons for inputs of size $n$. Hence, we primarily focus on designing \emph{sublinear} size networks -- with $n^{1-c}$ auxiliary neurons for some $c$.

\paragraph{Past Work: WTA}
Recently, we studied the `winner-take-all' (WTA) leader election task in SNNs \cite{LynchMP16}. Given a set of firing input neurons, the network is required to converge to a single firing output -- corresponding to the `winning' input.
In that work, we critically leveraged the noisy behavior of our spiking neuron model: randomness is key in breaking the symmetry between initially identical firing inputs.
  
\paragraph{This Paper: Similarity Testing and Compression}
In this paper, we study the role of randomness in a different setting: for similarity testing and compression. Consider the basic similarity testing problem: given $X_1,X_2 \in \{0,1\}^n$, we wish to distinguish the case when $X_1 = X_2$ from the case when the Hamming distance between the vectors is large -- i.e., $\ham(X_1,X_2) \ge \epsilon n$ for some parameter $\epsilon$. This problem can be solved very efficiently using randomness -- it suffices to sample $O(\log n /\epsilon)$ indices and compare $X_1$ and $X_2$ at these positions to distinguish the two cases with high probability. Beyond similarity testing, similar compression approaches using random input subsampling or hashing can lead to very efficient routines for a number of data processing tasks.

\subsection{A Neuro-RAM Unit}
 
To implement the randomized similarity testing approach described above, and to serve as a foundation for other random compression methods in spiking networks, 
we design a basic \emph{indexing module}, or random access memory, which we call a \emph{neuro-RAM}. This module solves:
\begin{definition}[Indexing]\label{def:index}
Given $X \in  \{0,1\}^n$ and $Y \in  \{0,1\}^{\log n}$ which is interpreted as an integer in $\{0,...,n-1\}$, the indexing problem is to output the value of the $Y^{th}$ bit of $X$\footnote{Here, and throughout, for simplicity we assume $n$ is a power of $2$ so $\log n$ is an interger.}.
\end{definition}
Our neuro-RAM uses a sublinear number of auxiliary neurons and solves indexing with high probability on any input. We focus on characterizing the trade-off between the convergence time and network size of the neuro-RAM, giving nearly matching upper and lower bounds.

Generally, our results show that a compressed representation (e.g., the index $Y$) can be used to access a much larger datastore (e.g., $X$), using a very compact neural network.
While binary indexing is not very `neural' we can imagine similar ideas extending to more natural coding schemes used, for example, for memory retrieval, scent recognition, or other tasks.

\paragraph{Relation to Prior Work}
Significant work has employed
 random synaptic connections between neurons -- e.g., the Johnson-Lindenstrauss compression results of \cite{allen2014sparse} and the work of Valiant \cite{valiant2000circuits}.  While it is reasonable to assume that the initial synapses are random, biological mechanisms for changing connectivity (functional plasticity) act over relatively large time frames and cannot provide a new random sample of the network for each new input. 
In contrast, stochastic spiking neurons do provide fresh randomness to each computation. In general, transforming of a network with $m$ possible random edges to a network with fixed edges and stochastic neurons requires $\Omega(m)$ auxiliary neurons and thus fails to fulfill our sublinearity goal, as there is typically at least one possible outgoing edge from each input. Our neuro-RAM can be thought of as improving the naive simulation -- by reading a random entry of an input, we simulate a random edge from the specified neuron. Beyond similarity testing, we outline how our result can be used to implement Johnson-Lindenstrauss compression similar to \cite{allen2014sparse} without assuming random connectivity.

\subsection{Our Contributions}
\subsubsection{Efficient Neuro-RAM Unit}

Our primary upper bound result is the following:
\begin{theorem}[$t$-round Neuro-RAM]\label{thm:introMain}
For every integer $t\leq \sqrt{n}$, there is a (\emph{recurrent}) SNN with $O(n/t)$ auxiliary neurons that solves the indexing problem in $t$ rounds with high probability. In particular, there exists a neuro-RAM unit that contains $O(\sqrt{n})$ auxiliary neurons and solves the indexing problem in $O(\sqrt{n})$ rounds.
\end{theorem}
Above, and throughout the paper `with high probability' or w.h.p. to denotes with probability at least $1-1/n^c$ for some constant $c$.
Theorem \ref{thm:introMain} is proven in Section \ref{sec:selection}.

\paragraph{Neuro-RAM Construction}
The main idea is to first `encode' the firing pattern of the input neurons $X$ into the potentials of $t$ neurons. These encoding neurons will spike with some probability dependent on their potential. 
However, simply recording the firing rates of the neurons to estimate this probability is too inefficient. Instead, we use a `successive decoding strategy', in which the firing rates of the encoding neurons are estimated at finer and finer levels of approximation, and adjusted through recurrent excitation or inhibition as  decoding progresses. The strategy converges in $O(n/t)$ rounds -- the smaller $t$ is the more information is contained in the potential of a single neuron, and the longer decoding takes.

Theorem \ref{thm:introMain} shows a significant separation between our networks and traditional feedforward circuits where significantly sublinear sized indexing units are not possible. 

\begin{fact}[See Lower Bounds in \cite{koiran1996vc}]\label{intro:LBFact}
A  circuit solving the indexing problem that consists of AND/OR gates connected in a feedforward manner requires $\Theta(n)$ gates. A feedforward circuit using linear threshold gates requires $\Theta(n/\log n)$ gates.
\end{fact}

We note, however, that our indexing mechanism 
\emph{does not exploit the randomness of the spiking neurons}, and in fact can also be implemented with deterministic linear threshold gates. Thus, the separation between Theorem \ref{thm:introMain} and Fact \ref{intro:LBFact} is entirely due to the recurrent (non-feedforward) layout of our network. Since any recurrent network using $O(m)$ neurons and converging in $t$ rounds can be `unrolled' into a feedforward circuit using $O(mt)$ neurons, Fact \ref{intro:LBFact} shows that the tradeoff between network size and runtime in Theorem \ref{thm:introMain} is optimal up to a $\log n$ factor, as long as we use our spiking neurons in a way that can also be implemented with deterministic  threshold gates. However, it does not rule out improvements using more sophisticated randomized strategies.

\subsubsection{Lower Bound for Neuro-RAM in Spiking Networks}

Surprisingly, we are able to show that despite the restricted way in which we use our spiking neuron model, significant improvements are not possible:
\begin{theorem}[Lower Bound for Neuro-RAM in SNNs]\label{thm:lbIntro} Any SNN that solves indexing in $t$ rounds with high probability in our model must use at least $\Omega\left (\frac{n}{t\log^2 n} \right)$ auxiliary neurons.
\end{theorem}

Theorem \ref{thm:lbIntro}, whose proof is in Section \ref{sec:lower}, shows that the tradeoff in Theorem \ref{thm:introMain} is within a $\log^2 n$ factor of optimal. It matches the lower bound of Fact \ref{intro:LBFact} for deterministic threshold gates up to a $\log n$ factor, showing that there is not a significant difference in the power of stochastic neurons and deterministic gates in solving indexing.

\paragraph{Reduction from SNNs to Deterministic Circuits}
We first argue that the output distribution of any SNN is identical to the output distribution of an algorithm that first chooses a deterministic threshold circuit from some distribution and then applies it to the input. This is a powerful observation as it lets us apply Yao's principle: an SNN lower bound can be shown via a lower bound for deterministic circuits on any input distribution \cite{yao1977probabilistic}.

\paragraph{Deterministic Circuit Lower Bound via VC Dimension}
We next show that any deterministic circuit that succeeds with high probability on the uniform input distribution cannot be too small. The bound is via a VC dimension-based argument, which extends the work of \cite{koiran1996vc} on indexing circuits.
As far as we are aware, we are the first to give a VC dimension-based lower bound for probabilistic and biologically plausible network architectures and we hope our work significantly expands the toolkit for proving lower bounds in this area. In contrast to our lower bounds on the WTA problem \cite{LynchMP16}, which rely on indistinguishability arguments based on network structure, our new techniques allow us to give more general bounds for any network architecture.  

\paragraph{Separation of Network Models}
Aside from showing that randomness does not give significant advantages in constructing a neuro-RAM (contrasting with its importance in WTA and similarity testing), our proof of Theorem \ref{thm:lbIntro} establishes a separation between feedforward spiking networks and deterministic \emph{sigmoidal circuits}. Our neurons spike with probability computed as a sigmoid of their membrane potential. In sigmoidal circuits, neurons output real numbers, equivalent to our spiking probabilities. A neuro-RAM can be implemented very efficiently in these networks:
\begin{fact}[See \cite{koiran1996vc}, along with \cite{maass1997networks} for similar bounds]\label{fact:sigmoid}
There is a feedforward sigmoidal circuit solving the indexing problem using $O(n^{1/2})$ gates.\footnote{Note that \cite{maass1991computational} shows that general deterministic sigmoidal circuits can be simulated by our spiking model. However, the simulation blows up the size of the circuit size by $\sqrt{n}$, giving $\Theta(n)$ auxiliary neurons.}
\end{fact}
In contrast, via an unrolling argument, the proof of Theorem \ref{thm:lbIntro} shows that any feedforward spiking network requires $\Omega \left ( \frac{n}{\log^2 n} \right )$ gates to solve indexing with high probability.

It has been shown that feedforward sigmoidal circuits can significantly outperform standard feedforward linear threshold circuits \cite{maass1991computational,koiran1996vc}. However, previously it was not known that restricting gates to spike with a sigmoid probability function rather than output the real value of this function significantly affected their power. Our lower bound, along with Fact \ref{fact:sigmoid}, shows that in some cases it does. This separation highlights the importance of modeling spiking neuron behavior in understanding complexity tradeoffs in neural computation.


\subsubsection{Applications to Randomized Similarity Testing and Compression}
As discussed, our neuro-RAM is widely applicable to algorithms that require random sampling of inputs. In Section \ref{sec:applications} we discuss our main application, to similarity testing -- i.e., testing if $X_1 = X_2$ or if $\ham(X_1,X_2) \ge \epsilon n$. 
It is easy to implement an exact equality tester using $\Theta(n)$ auxiliary neurons. Alternatively, one can solve exact equality with three auxiliary neurons using mixed positive and negative edge weights for the outgoing edges of inputs. However this is not biologically plausible -- neurons typically have either all positive (excitatory) or all negative (inhibitory) outgoing edges, a restriction included in our model.
Designing sublinear sized exact equality testers under this restriction seems difficult -- simulating the three neuron solution requires at least $\Theta(n)$ auxiliary neurons -- $\Theta(1)$ for each input.

By relaxing to similarity testing and applying our neuro-RAM, we can achieve sublinear sized networks. We can use $\Theta(\log n /\epsilon)$ neuro-RAMs, each with $O(\sqrt{n})$ auxiliary neurons to check equality at $\Theta(\log n /\epsilon)$ random positions of $X_1$ and $X_2$ distinguishing if $X_1 = X_2$ or if $\ham(X_1,X_2) \ge \epsilon n$ with high probability. This is the first sublinear solution for this problem in the spiking neural networks. In Section \ref{sec:applications}, we discuss possible additional applications of our neuro-RAM to Johnson-Lindenstrauss random compression, which amounts to multiplying the input by a sparse random matrix -- a generalization of input sampling.
%
%

\section{Computational Model and Preliminaries}\label{sec:model}
\subsection{Network Structure}
We now give a formal definition of our computational model.
A \emph{Spiking Neural Network} (SNN) $\Net =\langle X,Z,A, w,b \rangle$ consists of $n$ input neurons $X=\{x_1, \ldots, x_{n}\}$, $m$ output neurons $Z=\{z_1, \ldots, z_{m}\}$, and $\ell$ auxiliary neurons $A = \{a_1,...,a_{\ell} \}$. The directed, weighted synaptic connections between $X$, $Z$, and $A$ are described by the weight function $w: [X \cup Z \cup A] \times [X \cup Z \cup A] \rightarrow \mathbb{R}$. A weight $w(u,v) =0$ indicates that a connection is not present between neurons $u$ and $v$. Finally, for any neuron $v$, $\Bias(v) \in \mathbb{R}_{\geq 0}$ is the activation bias -- as we will see, roughly, $v$'s membrane potential must reach $\Bias(v)$ for a spike to occur with good probability.

The weight function defining the synapses in our networks is restricted in a few notable ways.
The in-degree of every input neuron $x_i$ is zero. That is, $w(u,x) = 0$ for all $u \in [X \cup Z \cup A]$ and $x \in X$. This restriction bears in mind 
that the input layer 
might in fact be the output layer of another network
and so incoming connections are avoided to allow for the composition of networks in higher level modular designs.
Additionally, each neuron is either inhibitory or excitatory: 
if $v$ is inhibitory, then $w(v,u)\leq 0$ for every $u$, and if $v$ is 
excitatory, then $w(v,u)\geq 0$ for every $u$. All input and output neurons are excitatory.

\subsection{Network Dynamics}
An SNN evolves in discrete, synchronous rounds as a Markov chain. 
The firing probability of every neuron at time $t$ depends on the firing status of its neighbors at time $t-1$, via a standard sigmoid 
function, with details given below.

For each neuron $u$, and each time $t \ge 0$, let $u^{t}=1$ if $u$ fires (i.e., generates a spike) at time $t$. Let $u^{0}$ denote the initial firing state of the neuron. Our results will specify the initial input firing states $x_j^{0} = 1$ and assume that $u^{0} = 0$ for all $u \in [Z \cup A]$.
For each non-input neuron $u$ and every $t \ge 1$, let $pot(u,t)$ denote the membrane potential at round $t$ and $p(u,t)$ denote
the corresponding firing probability ($\Pr[u^t = 1]$). These values are calculated as:
\begin{align}
\label{eq:potentialOut}
pot(u,t)=   \sum_{v \in X\cup Z \cup A}w_{v,u}\cdot v^{t-1} -b(u) 
\text{ and }p(u,t)=\frac{1}{1+e^{-pot(u,t)/\lambda}}
\end{align}
where $\lambda > 0$ is a \emph{temperature parameter}, which determines the steepness of the sigmoid. It is easy to see that $\lambda$ does not affect the computational power of the network. A network can be made to work with any $\lambda$ simply by scaling the synapse weights and biases appropriately.

For simplicity we assume throughout that $\lambda= \frac{1}{\Theta(\log n)}$. Thus by \eqref{eq:potentialOut}, if $pot(u,t) \ge 1$, then $u^t = 1$ \whp and if $pot(u,t) \le -1$, $u^t = 0$ \whp (recall that \whp denotes with probability at least $1-1/n^c$ for some constant $c$). Aside from this fact, the only other consequence of \eqref{eq:potentialOut} we use in our network constructions is that $pot(u,t) = 0 \implies p(u,t) = 1/2$. That is, we will use our spiking neurons entirely as random threshold gates, which fire \whp when the incoming potential from their neighbors' spikes exceeds $\Bias(u)$, don't fire \whp when the potential is below $\Bias(u)$, and fire randomly when the input potential equals the bias. It is an interesting open question if there are any problems which require using the full power of the sigmoidal probability function.

\subsection{Additional Notation}
For any vector $x$ we let $x_i$ denote the value at its $i^{th}$ position, starting from $x_0$.
Given binary $x \in \{0,1\}^n$, we use $\dec(x)$ to indicate the integer encoded by $x$. That is, $\dec(x) = \sum_{i=0}^{n-1} x_i \cdot 2^i$. Given an integer $x$ we use $\bin(x)$ to denote its binary encoding, where the number of digits used in the encoding will be clear from context. We will often think of the firing pattern of a set of neurons as a binary string. If $B = \{y_1,...,y_m\}$ is a set of $m$ neurons then $B^t \in \{0,1\}^m$ is the binary string corresponding to their firing pattern at time $t$. Since the input is typically fixed for some number of rounds, we often just write $X$ to refer to the $n$-bit string corresponding to the input firing pattern.

\paragraph{Boolean Circuits.}
We mention that SNNs are similar to boolean circuits,  
which have received enormous attention in theoretical computer science. 
A circuit consists of gates (e.g., threshold gates, probabilistic threshold gates) connected in a
directed acyclic graph. This restriction means that a circuit does not have 
 feedback connections or self-loops, which we do use in our SNNs. While we do not work with circuits directly, for our lower bound, we show a transformation from an SNN to a linear threshold circuit. We sometimes refer to circuits as \emph{feedforward} networks, indicating that their connections are cycle-free.
\section{Neuro-RAM Network}\label{sec:selection}
In this section we prove our main upper bound:
\begin{theorem}[Efficient Neuro-RAM Network]\label{thm:selection}
There exists an SSN with $O(\sqrt{n})$ auxiliary neurons that solves indexing in $5\sqrt{n}$ rounds. Specifically, given inputs $X \in \{0,1\}^n$, and $Y \in \{0,1 \}^{\log n}$, which are  fixed for all rounds $t \in \{0,...,5\sqrt{n}\}$, the output neuron $z$ satisfies: if $X_{\dec(Y)}=1$ then $z^{5\sqrt{n}} = 1$ \whp Otherwise, if $X_{\dec(Y)}=0$, $z^{5\sqrt{n}} = 0$ \whp
\end{theorem}

Theorem \ref{thm:selection} easily generalizes to other network sizes, giving Theorem \ref{thm:introMain}, which states the general size-time tradeoff.
Here we discuss the basic construction and intuition behind our network construction. The full details and proof are given in Appendice \ref{sec:bucket} and \ref{sec:decode}.
%

We divide the $n$ input neurons X into $\sqrt{n}$ buckets each containing $\sqrt{n}$ neurons\footnote{Throughout we assume for simplicity that $n = 2^{2m}$ for some integer $m$. This ensures that $\sqrt{n}$, $\log n$, and $\log \sqrt{n}$ are integers. It will be clear that if this is not the case, we can simply pad the input, which only affects our time and network size bounds by constant factors.}: 
$$X_0 = \{x_0,...,x_{\sqrt{n}-1}\},...,X_{\sqrt{n}-1} = \{x_{(\sqrt{n}-1)\sqrt{n}},...,x_{n-1}\}.$$
Throughout, all our indices start from $0$. 
We encode the firing pattern of each bucket $X_i$ via the potential of a \emph{single} neuron $e_i$. Set
$\set{w(x_{i\sqrt{n} + j},e_i) = 2^{\sqrt{n}-j}}$
for all $i,j\geq 0$. In this way, for every round $t$, the total potential contributed to $e_i$ by the firing of the inputs in bucket $X_i$ is equal to:
\begin{align}\label{encoding}
\sum_{j=0}^{\sqrt{n}-1} x_{i\sqrt{n} + j} \cdot 2^{\sqrt{n}-j} = 2\cdot\dec(\bar{X}_i).
\end{align}
where $\bar{X}_i$ is the reversal of $X_i$ and $\dec(\cdot)$ gives the decimal value of a binary string, as defined in the preliminaries.
\set{We set $\Bias(e_i) = 2^{\sqrt{n}+2} + 2^{\sqrt{n}} - 1$.} We will see later why this is an appropriate value.
We defer detailed discussion of the remaining connections to $e_i$ for now, first giving a general description of the network construction.

In addition to the \emph{encoding neurons} $e_0,...,e_{\sqrt{n}-1}$, we have \emph{decoding neurons} $d_{0,k},...,d_{\sqrt{n}-1,k}$ for $k = 1,2,3$ ($3\sqrt{n}$ neurons total). 
The idea is to select a bucket $X_i$ (via $e_i$) using the first $\log \sqrt{n} = \frac{\log n}{2}$ bits in the index $Y$. Let $Y_1 \eqdef \{y_0,...,y_{\frac{\log n}{2}-1}\}$ and $Y_2 \eqdef \{y_{\frac{\log n}{2}},...,y_{\log n-1} \}$ be the higher and lower order bits of $Y$ respectively. It is not hard to see that using $O(\sqrt{n})$ neurons we can construct a network that processes $Y_1$ and uses it to select $e_i$ with $i = \dec(Y_1)$. 
When a bucket is selected, the potential of any $e_j$ with $j \neq \dec(Y_1)$ is significantly depressed compared to that of $e_i$ and so after this selection stage, only $e_i$ fires.

We will then use the decoding neurons to `read' \emph{each bit of the potential encoded in $e_i$}. The final output is selected from each of these bits using the lower order bits $Y_2$, which can again be done efficiently with $O(\sqrt{n})$ neurons. We call this phase the decoding phase since the bucket neuron $e_i$ encodes the value (in decimal) of its bucket $X_i$, and we need to decode from that value the bit of the appropriate neuron inside that bucket. 

The decoding process works as follows: initially, $e_i$ will fire only if the \emph{first bit} of bucket $i$ is on. Note that the weight from this bit to $e_i$ is $2^{\sqrt{n}}$ and thus more than double the weight from any other input bit. Thus, by appropriately setting $\Bias(e_i)$, we can ensure that the setting of this single bit determines if $e_i$ fires initially.

\setlength{\columnsep}{17pt}%
\begin{wrapfigure}{r}{7cm}
\includegraphics[width=0.38\textwidth]{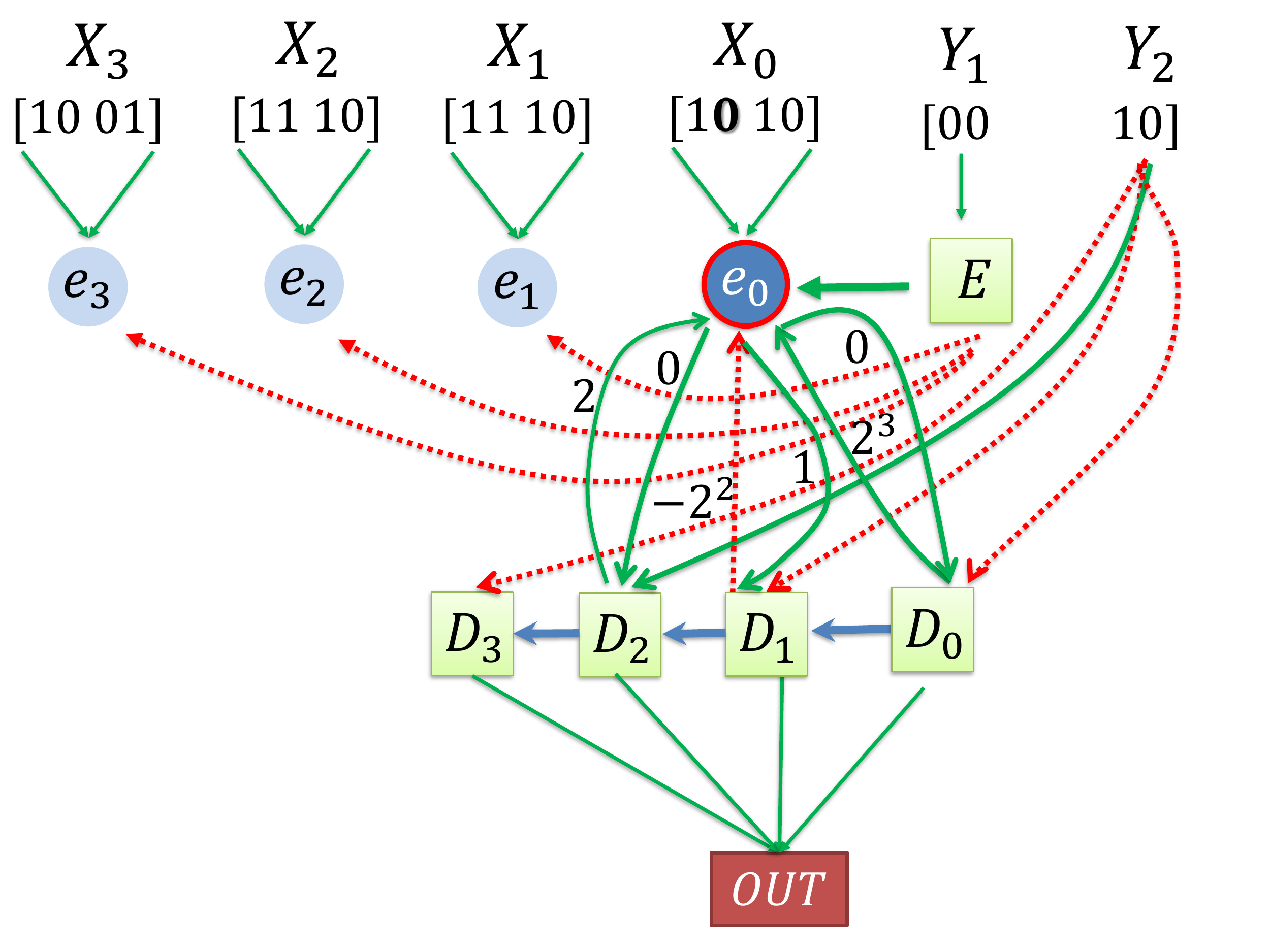}
	\caption{\footnotesize Illustration of the Neuro-RAM module. The index encoded by $Y$ is marked in bold.}
	\label{fig:neuroram}
\end{wrapfigure}
 If the first bit is the correct bit to output (i.e. if the last $\frac{\log n}{2}$ bits of the index $Y_2$ encode position $0$), this will trigger the output $z$ to fire. Otherwise, we iterate. If $e_i$ in fact fired, this triggers inhibition that cancels out the potential due to the first bit of bucket $i$. Thus, $e_i$ will now only fire if the \emph{second bit} of $X_i$ is on. If $e_i$ did not fire, the opposite will happen. Further excitation will be given to $e_i$ again ensuring that it can fire as long as the second bit of $X_i$ is on. The network iterates in this way, successively reading each bit, until we reach the one encoded by $Y_2$ and the output fires. The first decoding neuron for position $j$, $d_{j,1}$, is responsible to triggering the output to fire if $j$ is the correct bit encoded by $Y_2$. The second decoding neuron $d_{j,2}$ is responsible for providing excitation when $e_i$ does not fire. Finally, the third decoding neuron $d_{j,3}$ provides inhibition when $e_i$ does fire.

In Appendix \ref{sec:bucket}, we describe the first stage in which we use the first $\log n /2$ index bits to select the bucket to which the desired index belongs to. 

In Appendix \ref{sec:decode}, we discuss the second phase where we use the last $\log n/2$ bits of $Y$, to select the desired index inside the bucket $i$. Our success decoding process is synchronized by a clock mechanism, shown in Appendix \ref{sec:clock}. This clock mechanism consists of chain of $\Theta(\sqrt{n})$ neurons that govern the timing of the $\Theta(\sqrt{n})$ steps of our decoding scheme. Roughly speaking, traversing the $\sqrt{n}$ bits of the chosen $i^{th}$ bucket from left to right, we spend $O(1)$ rounds checking if the current index is the one encoded by $Y_2$.  If yes, we output the value at that index and if not, the clock will ``tick'' and we move to the next candidate. This successive decoding scheme is explained in Appendix \ref{sec:sucdecode}. 
\def\APENDSELECTION{
\subsection{First Stage: Bucket Selection}\label{sec:bucket}
To implement the bucket selection stage, we connect the neurons in $Y_1$ to each $e_i$ such that if 
$i = \dec(Y_1)$ the potential of $e_i$ is increased significantly, and if $i \neq \dec(Y_1)$ the potential of $e_i$ remains the same. By setting this potential increase to a very large value and the bias $\Bias(e_i)$ to a correspondingly very large value, $e_i$ will not fire with high probability unless $i = \dec(Y_1)$. This selection phase can be implemented with $O(\sqrt{n})$ auxilary neurons. 

Specifically, for each $y_j \in Y_1$, we have two neurons $y_{j,1}$, $y_{j,1'}$ connected as follows:

\begin{itemize}
\item $y_{j,1}$ is an excitatory neuron with \set{$\Bias(y_{j,1}) = 1$ and $w(y_j, y_{j,1}) = 2$}. In this way, if $y_j^t = 1$, $\pot(y_{j,1},t) = 1$ and so $y_{i,1}^{t+1} = 1$ \whp
\item $y_{j,1'}$ is an inhibitory neuron with \set{$\Bias(y_{j,1'}) = 1$ and $w(y_j, y_{j,1'}) = 2$}. So again, if $y_j^t = 1$, $\pot(y_{j,1'},t) = 1$ and so $y_{j,1}^{t+1} = 1$ \whp
\end{itemize}

The behavior of $y_{j,1}$ and $y_{j,1'}$ can be summarized as:
\begin{lemma}\label{lem:onoff1} For any $t$, if $y_j^t = 1$ then \whp $y_{j,1} = y_{j,1'} = 1$.
\end{lemma}

For each $e_i$, we then have an auxiliary excitatory neuron $g_i$. Connected as follows:

\begin{itemize}
\item \set{$w(y_{j,1},g_i) = 2$ if $\bin(i)_j = 1$ and $0$ otherwise}.
\item \set{$w(y_{j,1'},g_i) = -2$ if $\bin(i)_j = 0$ and $0$ otherwise}.
\item \set{$\Bias(g_i) = 2\norm{\bin(i)}_1-1$}.
\item \set{$w(g_i,e_i) = 2^{\sqrt{n}+2}$}.
\end{itemize}

We have the following lemma:
\begin{lemma}\label{lem:onoff2} For any  $t$, if $Y_1^t = \bin(i)$ then \whp $g_i^{t+2} = 1$ and $g_k^{t+2} = 0$ for all $k \neq i$.
\end{lemma}
\begin{proof}
The connections to $y_{j,1}$ and $y_{j,1'}$ and Lemma \ref{lem:onoff1} ensure that if $Y_1^t = \bin(i)$, \whp
\begin{align*}
\pot(g_i,t+1) &= -\Bias(g_j) + \sum_{j=0}^{\log n/2-1} \left [w(y_{j,1},g_i) \cdot y_{j,1}^{t+1} + w(y_{j,1'},g_i) \cdot y_{j,1'}^{t+1}\right ]\\
&= 1 - 2\norm{\bin(i)}_1 +  2\sum_{j=0}^{\log n/2-1} \left [\bin(i)_j \cdot y_j^t + (\bin(i)_j-1) \cdot y_j^t\right ]\\
& = 1 - 2\norm{\bin(i)}_1 + 2\sum_{j=0}^{\log n/2-1} \left [\bin(i)_j^2 + (\bin(i)_j-1) \cdot \bin(i)_j\right ]\\
& = 1 - 2\norm{\bin(i)}_1 + 2\norm{\bin(i)}_1 = 1.
\end{align*}
Thus, if $Y_1^t = \bin(i)$, $g_i^{t+2} = 1$ \whp
Similarly, if $Y_1^t = \bin(i)$ then for any $k \neq i$:
\begin{align*}
\pot(g_k,t+1) &= 1 - 2\norm{\bin(k)}_1 +  2\sum_{j=0}^{\log n/2-1} \left [\bin(i)_j \cdot \bin(k)_j + (\bin(i)_j-1) \cdot \bin(k)_j\right ]\\
&\le 1 - 2\norm{\bin(k)}_1 +  2\norm{\bin(k)}_1 - 2 \le -1
\end{align*}
where the bound that the sum is $\le 2\norm{\bin(k)}_1 - 2$ follows from the fact that $\bin(i)_j \neq \bin(k)_j$ for some $j$. Thus $g_k^{t+2} = 0$ \whp completing the lemma by a union bound over all $g_k$.
\end{proof}

By Lemma \ref{lem:onoff2} we can ensure that $e_i$ only fires if it corresponds to the bucket selected by $Y$ -- i.e., if $Y_1 = \bin(i)$. To do this we need one more fact, which will be clear after defining our full network:
\begin{fact}\label{weightFact}
The total weight of incoming excitatory connections to $e_i$ for any $i$, excluding the connection from $g_i$, is upper bounded by $2^{\sqrt{n}+2}$.
\end{fact}
\begin{lemma}[Bucket Selection]\label{lem:bucket}
For any  $t$, if $Y_1^t = \bin(i)$ then \whp $e_k^{t+3} = 0$ for all $k \neq i$.
\end{lemma}
\begin{proof}
By the setting of $\Bias(e_k) = 2^{\sqrt{n}+2} + 2^{\sqrt{n}}-1$, Fact \ref{weightFact}, and Lemma \ref{lem:onoff2} \whp
\begin{align*}
\pot(e_k,t+2) &\le - 2^{\sqrt{n}+2} - 2^{\sqrt{n}} + 1 + g_k^{t+2} \cdot w(g_i,e_i) + 2^{\sqrt{n}+2}\\
&\le -2^{\sqrt{n}} + 1 + 0  \le -1
\end{align*}
so \whp $e_k^{t+3} = 0$, giving the lemma after union bounding over all $e_k$.
\end{proof}
We note that if $Y_1^t = \bin(i)$ then $e_i$ will receive a potential of $2^{\sqrt{n}+2}$ from $g_i$ and so its effective bias will shift to $2^{\sqrt{n}}+1$, which as we will see, will be appropriate for the remainder of the algorithm.

\subsection{Second Stage: Bucket Decoding }\label{sec:decode}
We now discuss the decoding phase of our network.

\subsubsection{Clock Mechanism}\label{sec:clock}
For this phase we need a clock mechanism. Specifically, we have some initiator neuron $c_0$, with \set{$w(x_i,c_0) = 2$ for all inputs $x_i$ and $\Bias(c_0) = 1$}. Thus, $c_0$ fires \whp in response to at least one input firing. We then have $5\sqrt{n}$ additional excitatory neurons $c_1,...,c_{5\sqrt{n}}$ and $5\sqrt{n}$ additional inhibitory neurons $c_{1'},...,c_{5\sqrt{n}'}$. We set \set{$\Bias(c_i) = \Bias(c_{i'}) = 1$ and $w(c_{i-1},c_i) = w(c_{i-1},c_i')= 2$. Further, we set $w(c_1',c_1)= -2$, and for all $i < 5 \sqrt{n}$, $w(c_i', c_0) = -2n$}. This gives the following clocking property:
\begin{lemma}[Clock Mechanism]\label{lem:clock}
If $c_0^t = 1$ for some round $t$, then for all $i\le 5\sqrt{n}$, $c_i^{t+i} = 1$. Further, for all $j \le 5\sqrt{n}$, with $j \neq i$ $c_i^{t+j} = 0$ \whp
\end{lemma}
We place inhibitory connections from each $c_i'$ back to $c_0$, which prevent the clock from `restarting' before it has finished a complete cycle. We \emph{do not} connect the last inhibitor $c_{5\sqrt{n}}$ to $c_0$ (in fact, this inhibitor can be removed from the network). This ensures that once the $c_{5\sqrt{n}}$ fires, in the next round, as long as at least one input is active, $c_0$ will fire \whp restarting the clock. While this is not necessary for the correctness of our neuro-RAM, it will be useful in applications that reuse this network to process multiple inputs.
\begin{proof}
Since $c_0^t = 1$, $\pot(c_1,t) = \pot(c_{1'},t)  = - 1 + 2 = 1$ and so $c_1^{t+1} = c_{1'}^{t+1} = 1$ \whp Thus, due to the inhibitory connection from $c_{1'}$, and since $c_0$ has a total of $2n$ excitatory weight from all $n$ inputs, \whp $\pot(c_0,t+1) \le -1 + 2n - 2n \le -1$ so $c_0^{t+2} = 0$ \whp Finally, due to the inhibitory connection from $c_{1'}$ to $c_{1}$, \whp $\pot(c_{1},t+1) \le -1 + 2 -2 \le -1$ so $c_{1}^{t+2} = 0$ \whp

Now for any $i > 1$ assume by induction that for $j < i$, $c_j^{t+j}=1$, $c_j^{t+j+1} = 0$ and $c_0^{t+j+1}$ = 0. This implies that $\pot(c_i,t+i-1) = \pot(c_{i'},t+i-1) = 1$ 
and thus $c_i^{t+i} = c_{i'}^{t+i} = 1$ \whp Additionally, since $c_{i-1}^{t+i} = 0$, $\pot(c_i,t+i) = -1$ and so $c_i^{t+i+1} = 0$ \whp Finally, since $c_{i'}^{t+i} = 1$ \whp $\pot(c_0,t+i) \le -1 + 2n - 2n \le -1$ and so $c_0^{t+i+1} = 0$ \whp Thus, the inductive assumption is fulfilled up to round $i$.

The lemma follows simply by noting that since $c_0$ does not fire in any round after round $t+1$, it never excites $c_1$ and so later neurons in the chain are not re-excited and at round $t+i$, $c_j^{t+i} = 0 $ \whp for all $j \le i-1$. We union bound over all failure events and have the full result \whp
\end{proof}

We finally state the following immediate corollary of Lemma \ref{lem:clock} which will be useful:
\begin{corollary}\label{cor:clock}
If any input in $X$ fires at time $t$ then \whp for all $i$, $c_i^{i+1} = 1$ and for all $j \le 5\sqrt{n}$ with $j \neq i$, $c_i^{j+1} = 0$.
\end{corollary}
\begin{proof}
This just follows from the fact that if any input fires in round $t$, $c_0$ fires \whp in round $t+1$. We then have the result by Lemma \ref{lem:clock}.
\end{proof}

\subsubsection{Successive Decoding}\label{sec:sucdecode}

With the clock mechanism of Lemma \ref{lem:clock} in place we can give our decoding algorithm. 
As discussed,
we have three decoding neurons for each bit, labeled $d_{j,1}$, $d_{j,2}$, and $d_{j,3}$. 
We describe the specific setups for each below.

\paragraph{Output Triggering}
$d_{j,1}$ is responsible for triggering the output $z$ to fire if bit $j$ of $X_{\dec(Y_1)}$ is $1$. We set \set{$w(e_i,d_{j,1}) = 2$ for all $i,j$.
We additionally set $w(d_{j,1},z) = 2$ for all $j$ and $\Bias(z) = 1$}, such that whenever one of these triggering neurons fires, the output $z$ fires. We also add a self-loop with weight $w(z,z) = 2$ such that once the output fires, it continues to fire \whp

 Like $e_i$, $d_{j,1}$ is equipped with an auxilary neuron $f_j$ which fires \whp iff $Y_2 = \bin(j)$. Specifically, we connect $f_j$ to $Y_2$ in an identical manner to how we connected $g_j$ to $Y_1$ and have the following analog of Lemma \ref{lem:onoff2}:
\begin{lemma}\label{lem:onoff3} For any  $t$, if $Y_2^t = \bin(j)$ then \whp $f_j^{t+2} = 1$ and $f_k^{t+2} = 0$ for all $k \neq j$.
\end{lemma}
We set \set{$w(y_j, d_{j,1}) = 2$}. Additionally, to insure that $d_{j,1}$ only fires at the appropriate step of decoding we connect it to our clock mechanism. We set \set{$w(c_l, d_{j,1}) = 2\sqrt{n}$ for $l = 5j+2$. Finally we set $\Bias(d_{j,1}) = 2\sqrt{n} + 3$}. This gives the following lemma:
\begin{lemma}\label{triggering} Assume that the inputs $X$, $Y_1$, and $Y_2$ remain fixed for $t \in \{0,...,5\sqrt{n}\}$ and that $Y_1 = \bin(i)$ and $Y_2 = \bin(j)$. Then
$d_{j,1}^t = 1$ for $t = 5j+4$ \whp if $e_i^{t} = 1$ for $t = 5j+3$. Otherwise, \whp $d_{j,1}^t = 0$ for all $t \le 5\sqrt{n}$.
\end{lemma}
\begin{proof}
$d_{j,1}$ will not fire \whp in round $t$ unless $c_l$ (for $l = 5j+2$) fires in round $t-1$. This is because the weight of all incoming connections to $d_{j,1}$ from the $\sqrt{n}$ encoding neurons and $y_j$ is $2\sqrt{n} + 2$ which is not enough to overcome the bias $\Bias(d_{j,1}) = 2\sqrt{n}+3$. 

Additionally, assuming $X^t \neq 0$, then at least one input fires in round $0$, so by Corollary \ref{cor:clock}, $c_l$ fires \whp in round $5j+ 3$ and in no other rounds by Lemma \ref{lem:clock}. So if $d_{j,1}$ fires, \whp it must be in round $t = 5j+ 4$.

Note that to have $d_{j,1}^{t} = 1$ \whp  we must additionally have $f_j^{t-1} = 1$ and $e_{\dec(Y_1)}^{t-1} = 1$.
Otherwise, since by Lemma \ref{lem:bucket}, $e_{\dec(Y_1)}$ is the only encoding neuron that fires after round $3$:
$$\pot(d_{j,1},t-1) \le -\Bias(d_{j,1}) + w(e_{\dec(Y_1)},d_{j,1}) + w(c_l,d_{j,1}) \le -2\sqrt{n} - 3  + 2 + 2\sqrt{n} \le -1$$
 and so $d_{j,1}^{t} = 0$ \whp
By Lemma \ref{lem:onoff3}, if $Y_2^{t'} = \bin(j)$ for all $t' \in \{0,...,5\sqrt{n}\}$ then $y_j$ will fire in all rounds after round $2$ and hence in round $t-1$. Thus $d_{1,j}$ will fire \whp if $e_{\dec{Y_1}}$ fires in $t-1$ as well. If $Y_2 \neq \bin(j)$ or if $e_{\dec{Y_1}}$ does not fire in this round, then $d_{1,j}$ will not fire with high probability, giving the lemma. We conclude by noting that we assumed that $X^t \neq 0$. If $X^t = 0$, then $c_0$ will never be triggered, and thus no $d_{j,1}$ will ever fire, so $z$ will never fire. This is a correct output, as all bits of the input are $0$.
 \end{proof}
 
 From Lemma \ref{triggering} we have the following simple corollary:
 \begin{corollary}\label{cor:trigger}
 Assume that the inputs $X$, $Y_1$, and $Y_2$ remain fixed for $t \in \{0,...,5\sqrt{n}\}$ and that $Y_1 = \bin(i)$ and $Y_2 = \bin(j)$. Then
$z^{5\sqrt{n}} = 1$ \whp if $e_i^{t} = 1$ for $t = 5j+3$. $z^{5\sqrt{n}} = 0$ \whp otherwise.
\end{corollary}
\begin{proof}
This follows directly from Lemma \ref{triggering} and the fact that we set $\Bias(z) = 1$ and $w(d_{j,1},z) = 2$ for all $j \in 1,...,\sqrt{n}$. Additionally, once $z$ fires, it continues firing as we added a self-loop with weight $2$. Thus it fires \whp in round $5\sqrt{n}$.
\end{proof}
 
 \paragraph{Potential Reading}
  Corollary \ref{cor:trigger} shows that as long as $\bin(i) = Y_1$, $\bin(j) = Y_2$, and $X_{\sqrt{n} i + j}= 1$ causes $e_i$ to fire at round $5j+3$ then $z$ will fire. If $e_i$ does not fire in this round then $z$ will fire in a round \whp iff $X_{\dec(Y)} = 1$. Thus is remains to demonstrate how to ensure that $e_i$ fires in this round if $X_{\sqrt{n} i + j}= 1$ and does not fire if $X_{\sqrt{n} i + j}= 0$.

To do this we use an excitatory neurons $d_{j,2},d_{j,3'}$ and the inhibitory neuron $d_{j,3}$. We set \set{$\Bias(d_{j,2}) = 1$ and $\Bias(d_{j,3}) = \Bias(d_{j,3'}) = 3$. We set $w(e_i,d_{j,3}) = w(e_i,d_{j,3'}) =2$ for all $i,j \in \{0,...,\sqrt{n}-1\}$} and finally \set{$w(c_l,d_{j,2}) = 2$ for $l = 5j+2$ and $w(c_l,d_{j,3}) = 2$ for $l = 5j+2$. Finally we create self loops $w(d_{j,2},d_{j,2}) = 2$ and $w(d_{j,3'},d_{j,3'}) = 4$.} and $w(d_{j,3'},d_{j,3}) = 4$.
This gives the following lemmas:

\begin{lemma}\label{d2response}Assume that $X^0 \neq 0$ (i.e. at least one input fires in round $0$). With high probability, for every $j$, $d_{j,2}^t = 0$ for all $t < 5j+4$ and 
$d_{j,2}^t = 1$ for all $5j+4 \le t \le 5\sqrt{n}$.
\end{lemma}
\begin{proof}
We have $\Bias(d_{j,2}) =1$, and $w(c_l, d_{j,2}) = 2$ so as long as $X^t \neq 0$, by Corollary \ref{cor:clock}, $c_l$ fires in round $5j+3$, $d_{j,2}$ will fire in the next round \whp Further, it will continue to fire in all successive rounds \whp due to its self-loop with $w(d_{j,2},d_{j,2}) = 2$. Since it does not fire in round $0$, its self-loop will be inactive and it will not fire in any round before $5j+4$.
\end{proof}

\begin{lemma}\label{d3response} 
Assume that $X$ and $Y_1$ remain fixed for $t \in \{0,...,5\sqrt{n}\}$, that $X\neq 0$, that $Y_1 = \bin(i)$.
For every $j$, if $e_i^{t} = 1$ for $t = 5j+3$, then \whp 
$d_{j,3}^t = 1$ for all $t \ge 5j+4$. Otherwise, \whp $d_{j,3}^t = 0$ for all $t \le 5\sqrt{n}$.
\end{lemma}
\begin{proof}
Since $\Bias(d_{j,3}) = 3$, in order of $d_{j,3}$ to fire in round $t$ if it did not fire in round $t-1$ (and hence $d_{j,3'}$, which has identical connections, was not activated \whp) we must have $e_i^{t-1} =1$ for some $i$ and $c_l^{t-1} = 1$ for $l = 5j+2$. By Corollary \ref{cor:clock}, since $X^0 \neq 0$, $c_l$ fires \whp in round $5j+3$ and no other round, so if $d_{j,3}$ fires it must be in round $5j+4$ \whp Since only $e_{\dec(Y_1)}$ fires in any round after round $3$ \whp by Lemma \ref{lem:bucket}, we must have $e_{\dec(Y_1)}^{5j+3} = 1$ in order for $d_{j,3}$ to fire. We finally note that once $d_{j,3}$ fires, it will continue firing \whp in each round due to $d_{j,3'}$ whose excitatory connection excites it (the connection from $d_{j,3'}$ behaves as an excitatory self-loop, which $d_{j,3}$ is not allowed to have directly since it is an inhibitor). Further, it does not fire in any round before $t = 5j+4$ since it does not fire in round $0$ so $d_{j,3'}$ will be inactive.
\end{proof}

We now discuss how $d_{j,2}$ and $d_{j,3}$ provide feedback to $e_i$. We set for all $i$, 
$$
\set{w(d_{j,2},e_i) = 2^{\sqrt{n}-j-1} \text{ and } w(d_{j,3},e_i) = -2^{\sqrt{n}-j} }.
$$
We can verify Fact \ref{weightFact}: The total weight on each $e_i$ from the $d_{j,2}$ neurons is at most $\sum_{j=0}^{\sqrt{n}-1} 2^{\sqrt{n}-j-1} \le 2^{\sqrt{n}}$. The total weight from the inputs is at most $\sum_{j=0}^{\sqrt{n}-1} 2^{\sqrt{n}-j} \le 2^{\sqrt{n}+1}$, so overall the total excitatory weight is at most $2^{\sqrt{n}+2}$. We can finally prove:
\begin{lemma}\label{mainLemma}
Assume that $X$, $Y_1$ remain fixed for $t \in \{0,...,5\sqrt{n}\}$ and that $Y_1 = \bin(i)$.
For all $k,j$,
$e_k^{5j+3} = 1$ \whp if $X_{k\sqrt{n} + j} = 1$ \emph{and} $k=i$. Otherwise, $e_k^{5j+3} = 0$ \whp
\end{lemma}
\begin{proof}
We first note that if $X = 0$, then no $e_k$ will ever fire and the lemma will be correct trivially. So we assume $X \neq 0$.
By Lemma \ref{lem:bucket}, for $k \neq i$, $e_k^{t}$ for all $3 \le t \le 5\sqrt{n}$ \whp which immediately gives the result in this case. So now consider $k = i$.
For $j = 1$, since we set $\Bias(e_i) = 2^{\sqrt{n}+2} + 2^{\sqrt{n}} -1$, since $Y_1^t = \bin(i)$ for $t = 0$, by Lemma \ref{lem:onoff2}, $g_i$ fires in round $2$. Thus \whp
\begin{align*}
\pot(e_i,2) &= -2^{\sqrt{n}+2} - 2^{\sqrt{n}} + 1 + w(g_i,e_i) + \sum_{j=0}^{\sqrt{n}-1} \left [d_{j,2}^{2} w(d_{j,2},e_i) + d_{j,3}^{2} w(d_{j,3},e_i)  \right ]\\ &+\sum_{j=0}^{\sqrt{n}-1} X_{i\sqrt{n} + j} \cdot 2^{\sqrt{n}-j}\\
&= - 2^{\sqrt{n}} + 1 + \sum_{j=0}^{\sqrt{n}-1} X_{i\sqrt{n} + j} \cdot 2^{\sqrt{n}-j}
\end{align*}
where the last step follows since $w(g_i,e_i) = 2^{\sqrt{n}+2}$ and since neither $d_{j,2}$ nor $d_{j,3}$ fire \whp  before round $4$ (see Lemmas \ref{d2response} and \ref{d3response}). Now, $\sum_{j=0}^{\sqrt{n}-1} X_{i\sqrt{n} + j} \cdot 2^{\sqrt{n}-j} \ge 2^{\sqrt{n}}$ if $X_{i\sqrt{n}} = 1$ and $\sum_{j=0}^{\sqrt{n}-1} X_{i\sqrt{n} + j} \cdot 2^{\sqrt{n}-j} \le 2^{\sqrt{n}} -2$ otherwise. Thus $e_i$ fires \whp in round $3$ if $X_{i\sqrt{n}} = 1$ and does not fire \whp otherwise.
This completes the lemma in the base case of $j=0$. 

Now consider $j \ge 1$ and assume the lemma holds \whp for all $j' < j$. By this inductive assumption and Lemmas \ref{d2response} and \ref{d3response} at round $t = 5j+2$ we have \whp
\begin{align*}
\pot(e_i,t) &= -2^{\sqrt{n+2}} - 2^{\sqrt{n}} + 1 + w(g_i,e_i) + \sum_{j=0}^{\sqrt{n}-1} \left [d_{j,2}^{t} w(d_{j,2},e_i) + d_{j,3}^{t} w(d_{j,3},e_i)  \right ] \\ &+ \sum_{j=0}^{\sqrt{n}-1} X_{i\sqrt{n} + j} \cdot 2^{\sqrt{n}-j}\\
&= 1 - 2^{\sqrt{n}} + \sum_{j'=1}^{j-1}  \left [2^{\sqrt{n}-j'-1} - X_{i\sqrt{n}+j'} 2^{\sqrt{n}-j}  \right ] + \sum_{j=0}^{\sqrt{n}-1} X_{i\sqrt{n} + j} \cdot 2^{\sqrt{n}-j}\\
&= 1 - 2^{\sqrt{n}} + \sum_{j'=0}^{j-1} 2^{\sqrt{n}-j'-1} + \sum_{j'=j}^{\sqrt{n}-1}  X_{i\sqrt{n}+j'} 2^{\sqrt{n}-j}\\
&= 1 - 2^{\sqrt{n}-j} + \sum_{j'=j}^{\sqrt{n}-1}  X_{i\sqrt{n}+j'} 2^{\sqrt{n}-j}
\end{align*}
And again, this potential is $\ge 1$ if $ X_{i\sqrt{n}+j} = 1$ and $\le -1$ otherwise. Thus, $e_i^{5j+3} = 1$ \whp if $X_{i\sqrt{n}+j} = 1$ and is $0$ otherwise \whp This completes the lemma.
\end{proof}

With this lemma in hand, we can now prove our main result Theorem \ref{thm:selection}.
\begin{proof}[Proof of Theorem \ref{thm:selection}]
The theorem follows from Corollary \ref{cor:trigger} combined with Lemma \ref{mainLemma}.
\end{proof}
}

Note that our model and the proof of Theorem \ref{thm:selection} assume that no auxiliary neurons or the output neuron fire in round $0$. 
However, in applications it will often be desirable to run the Neuro-RAM for multiple inputs, with execution not necessarily starting at round $0$. We can easily add a mechanism that `clears' the network once it outputs, giving:
\begin{observation}[Running Neuro-RAM for Multiple Inputs]\label{obs:multiple}
The Neuro-RAM of Theorem \ref{thm:selection} can be made to run correctly given a sequence of multiple inputs.
\end{observation}
\def\APPENDMANYA{
\begin{proof}[Proof of Observation \ref{obs:multiple}]
To run the indexing algorithm multiple times, the network must `reset' to a state in which all auxiliary neurons do not fire for a round. This is notably important for $d_{j,2}$, $d_{j,3'}$, and $z$. Once these neurons spike, they propagate the spike via a self-loop and will fire continuously \whp unless they receive some external inhibition. A simple way to achieve a reset is to add an inhibitory neuron $r$ with $\Bias(r) = 1$ and $w(c_{5\sqrt{n}-2},r) = 2$. Thus, since $c_{5\sqrt{n}-2}$ fires \whp in round $5\sqrt{n}-1$, $r$ will fire \whp in round $5\sqrt{n}$. We can add an inhibitory synapse from $r$ to each decoding neuron and to $z$, with arbitrarily large weight. In this way, in round $5\sqrt{n}+1$, none of these neurons will fire with high probability, and the computation will proceed as described. Round $5\sqrt{n}+1$ can be identified with round $0$ in the statement of Theorem \ref{thm:selection}. 
\end{proof}
}

\section{Lower Bound for Neuro-RAM in Spiking Networks}\label{sec:lower}
In this section, we show that our neuro-RAM construction is nearly optimal. Specifically:
\begin{theorem}\label{thm:lower}
Any SNN solving indexing with probability $\ge 1- \frac{1}{2n}$ in $t$ rounds must use $\ell = \Omega \left ( \frac{n}{t \log^2 n} \right )$ auxiliary neurons.
\end{theorem}
This result matches the lower bound for deterministic threshold gates of Fact \ref{intro:LBFact} up to a $\log n$ factor, demonstrating that the use of randomness cannot give significant runtime advantages for the indexing problem.
We note that even if one just desires a constant (e.g. $2/3$) probability of success -- the lower bound applies. By replicating any network with success probability $2/3$, $\Theta(\log n)$ times and taking the majority output (which can be computed with just a single additional auxiliary neuron), we obtain a network that solves the problem \whp We thus have:
\begin{corollary}\label{thm:lowerConstant}
Any SNN solving indexing with probability $\ge 2/3$ in $t$ rounds must use $\ell = \Omega \left ( \frac{n}{t \log^3 n} \right )$ auxiliary neurons.
\end{corollary}
The proof of Theorem \ref{thm:lower} proceeds in a number of steps, which we overview here.

\subsection{High Level Approach and Intuition}
\paragraph{Reduction to Deterministic Indexing Circuit.}
We first observe that a network with $\ell$ auxiliary neurons solving the indexing problem in $t$ rounds can be unrolled into a feedforward circuit with $t$ layers and $\ell$ neurons per layer. 
We then show that the output distribution of a feedforward stochastic spiking circuit is identical to the output distribution if we first draw a deterministic linear threshold circuit (still with $t$ layers and $\ell$ neurons per layer) from a certain distribution, and evaluate our input using this random circuit. 

This equivalence is powerful since it allows us to apply Yao's minimax principal \cite{yao1977probabilistic}: assuming the existence of a feedforward SNN solving indexing with probability $\ge 1-\frac{1}{2n}$, given any distribution of the inputs $X,Y$, there must be some deterministic linear threshold circuit $\mathcal{N}_D$ which solves indexing with probability $\ge 1-\frac{1}{2n}$ over this distribution.

If we consider the uniform distribution over $X,Y$, 
this success probablity ensures via an averaging argument that for at least $1/2$ of the $2^n$ possible values of $X$, $\mathcal{N}_D$ succeeds for at least a $1-\frac{1}{2n}$ fraction of the possible $Y$ inputs. Note, however, that the $Y$ can only take on $n$ possible values -- thus this ensures that for $1/2$ the possible values of $X$, $\mathcal{N}_D$ succeeds for \emph{all possible values of the index $Y$}. Let $\mathcal{X}$ be the set of `good inputs' for which $\mathcal{N}_D$ succeeds.

\paragraph{Lower Bound for Deterministic Indexing on a Subset of Inputs.}
We have now reduced our problem to giving a lower bound on the size of a deterministic linear threshold circuit which solves indexing on an arbitrary subset $\mathcal{X}$  of $\frac{1}{2} \cdot 2^n = 2^{n-1}$ inputs. We do this using VC dimension techniques inspired by the indexing lower bound of \cite{koiran1996vc}.

The key idea is to observe that if we fix some input $X \in \mathcal{X}$, then given $Y$, $\mathcal{N}_D$ evaluates the function $f_{X}: \{0,1\}^{\log n} \rightarrow \{0,1\}$, whose truth table is given by $X$. Thus $\mathcal{N}_D$ can be viewed as a circuit for evaluating any function $f_X(Y)$ for $X  \in \mathcal{X}$, where the $X$ inputs are `programmable parameters', which effectively change the thresholds of some gates. 

It can be shown that the VC dimension of the class of functions computable by a fixed a linear threshold circuit with $m$ gates and variable thresholds is $O(m \log m)$. Thus for a circuit with $t$ layers and $\ell$ gates per layer, the VC dimension is $O(\ell t \log(\ell t))$ \cite{baum1989size}.
Further, as a consequence of
Sauer's Lemma \cite{sauer1972density,shelah1972combinatorial,anthony2009neural}, defining the class of functions $\mathcal{F} = \{f_X\text{ for any } X \in \mathcal{X}\}$, since $|\mathcal{F}| = |\mathcal{X}| = 2^{n-1}$, we have $VC(\mathcal{F}) = \Theta(n/\log n)$. These two VC dimension bounds, in combination with the fact that we know $\mathcal{N}_D$ can compute any function in $\mathcal{F}$ if its input bits are fixed appropriately,
 imply that $\ell t \cdot \log(\ell t) = \Omega(n/\log n)$. Rearranging gives $\ell = \Omega \left ( \frac{n}{t \log^2n} \right)$, completing Theorem \ref{thm:lower}.

\subsection{Reduction to Deterministic Indexing Circuit}\label{lem:ff}
We now give the argument explained above in detail, first describing how any SNN that solves indexing \whp implies the existence of a deterministic feedforward linear threshold circuit which solves indexing for a large fraction of possible inputs $X$.
\begin{lemma}[Conversion to Feedforward Network]
Consider any SNN $\mathcal{N}$ with $\ell$ auxiliary neurons, which given input $X \in \{0,1\}^n$ that is fixed for rounds $\{0,...,t\}$, has output $z$ satisfying $\Pr[z^t = 1] = p$. Then there is a feedforward SNN $\mathcal{N}_F$ (an SNN whose directed edges form an acyclic graph) with $(t-1)\cdot (\ell+1)$ auxiliary neurons also satisfying $\Pr[z^t = 1 ] = p$ when given $X$ which is fixed for rounds $\{0,...,t\}$.
\end{lemma}
\begin{proof}
Let $B  = A \cup z$ -- all non-input neurons.
We simply produce $t-1$ duplicates of each auxiliary neuron $a \in A: \{a_1,...,a_{t-1}\}$ and of $z: \{z_1,...,z_{t-1}\}$, which are split into layers $B_1,...,B_{t-1}$. For each incoming edge from a neuron $u$ to $v$ and each $i \ge 2$ we add an identical edge from $u_{i-1}$ to $v_i$. Any incoming edges from input neurons to $u$ are added to each $u_i$ for all $i \ge 1$. Finally connect $z$ to the appropriate neurons in $B_{t-1}$ (which may  include $z_{t-1}$ if there is a self-loop in $\Net$).

In round $1$, the joint distribution of the spikes $B_1^1$ in $\mathcal{N}_F$ is identical to the distribution of $B^1$ in $\mathcal{N}$ since these neurons have identical incoming connections from the inputs, and since any incoming connections from other auxiliary neurons are not triggered in $\mathcal{N}$ since none of these neurons fire at time $0$.

Assuming via induction that $B_{i}^{i}$ is identically distributed to $B^{i}$, since $B_{i+1}$ only has incoming connections from $B_i$ and the inputs which are fixed, then the distribution of $B_{i+1}^{i+1}$ identical to that of $B^{i+1}$. Thus $B_{t-1}^{t-1}$ is identically distributed to $B^{t-1}$, and since the output in $\mathcal{N}_F$ is only connected to $B_{t-1}$ its distribution is the same in round $t$ as in $\mathcal{N}$.
\end{proof}

\begin{lemma}[Conversion to Distribution over Deterministic Threshold Circuits]\label{lem:det}
Consider any spiking sigmoidal network $\mathcal{N}$ with $\ell$ auxiliary neurons, which given input $X \in \{0,1\}^n$ that is fixed for rounds $\{0,...,t\}$, has output neuron $z$ satisfying $\Pr[z^t = 1] = p$. Then there is a distribution $\mathcal{D}$ over feedforward deterministic threshold circuits with $(t-1)\cdot (\ell +1)$ auxiliary gates that, for $\mathcal{N}_D \sim \mathcal{D}$ with output $z$,  $\Pr_{\mathcal{D}}[z^t = 1 ] = p$ when presented input $X$.
\end{lemma}
\begin{proof}
We start with $\mathcal{N}_F$ obtained from Lemma \ref{lem:ff}. This circuit has $t-1$ layers of $\ell+1$ neurons $B_1,...,B_{t-1}$. Given $X \in \{0,1\}^n$ that is fixed for rounds $\{0,...,t\}$, $\mathcal{N}_F$ has $\Pr[z^t = 1] = p$, which matches the firing probability of the output $z$ in $\mathcal{N}$ in round $t$.

Let $\mathcal{D}$ be a distribution on deterministic threshold circuits that have identical edge weights to $\mathcal{N}_F$. Additionally, for any (non-input) neuron $u \in \mathcal{N}_F$, letting $\bar{u}$ be the corresponding neuron in the deterministic circuit, set the bias $b(\bar{u}) = \eta$, where $\eta$ is distributed according to a logistic distribution with mean $\mu = b(u)$ and scale $s = \lambda$. The random bias is chosen independently for each $u$. It is well known that the cumulative density function of this distribution is equal to the sigmoid function. That is:
\begin{align}\label{eq:cdf}
\Pr[\eta \le x] &= \frac{1}{1+e^{-\frac{x - b(u)}{\lambda}}}.
\end{align}

Consider $\mathcal{N}_D \sim \mathcal{D}$ and any neuron $u$ in the first layer $B_1$ of $\mathcal{N}_F$. $u$ only has incoming edges from the input neurons $X$. Thus, its corresponding neuron $\bar{u}$ in $\mathcal{N}_D$ also only has incoming edges from the input neurons. Let $W = \sum_{x \in X} w(x,u) \cdot x^0$. Then we have:
\begin{align*}
\Pr_{\mathcal{D}} [\bar{u}^1 = 1] = \Pr [ W - \eta \ge 0] &= \Pr [\eta \ge W]\tag{Deterministic threshold}\\
&= \frac{1}{1+e^{-\frac{W-b(u)}{\lambda}}}\tag{Logistic distribution CDF \eqref{eq:cdf}}\\
&= \Pr [u^1 = 1].\tag{Spiking sigmoid dynamics \eqref{eq:potentialOut}}
\end{align*}
Let $\bar{B}_i$ denote the neurons in $\mathcal{N}_D$ corresponding to those in $B_i$.
Since in round $1$, all neurons in $B_1$ fire independently and since all neurons in $\bar{B}_1$ fire independently as their random biases are chosen independently, the joint firing distribution of $B_1^1$ is identical to that of $\bar{B}_1^1$.

By induction assume that $\bar{B}_i^i$ is identically distributed (over the random choice of deterministic network $\mathcal{N}_D \sim \mathcal{D}$) to $B_i^i$. Then for any $u \in B_{i+1}$ we have by the same argument as above, conditioning on some fixed firing pattern $V$ of $B_i$ in round $i$: 
$$
\Pr_{\mathcal{D}} [\bar{u}^{i+1} = 1 | \bar{B}_i^i = V] = \Pr [u^{i+1} = 1 | B_i^i = V].
$$
Conditioned on $B_i^i = V$, the neurons in $B_{i+1}$ fire independently in round $i+1$. So do the neurons of $\bar{B}_{i+1}$ due to their independent choices of random biases. Thus, the above implies that the distribution of $\bar{B}_{i+1}^{i+1}$ conditioned on $\bar{B}_i^i = V$ is identical to the distribution of $B_{i+1}^{i+1}$. This holds for all $V$, so, the full joint distribution of $\bar{B}_{i+1}^{i+1}$ is identical to that of ${B}_{i+1}^{i+1}$.

We conclude by noting that the same argument applies for the outputs of $\mathcal{N}_F$ and $\mathcal{N}_D$ since $\bar{B}_{t-1}^{t-1}$ is identically distributed to ${B}_{t-1}^{t-1}$. 
\end{proof}

Lemma \ref{lem:det} is simple but powerful -- it demonstrates the following:

\noindent
\begin{minipage}{1.02\textwidth}
\vspace{.5em}
\begin{mdframed}[hidealllines=false,backgroundcolor=gray!30]
The output distribution of a spiking sigmoid network is identical to the output distribution of a \emph{deterministic} feedforward threshold circuit drawn from some distribution $\mathcal{D}$.
\end{mdframed}
\vspace{.5em}
\end{minipage}

\noindent
Thus, the performance of any spiking sigmoid network is equivalent to the performance of a randomized algorithm which first selects a linear threshold circuit using $\mathcal{D}$ and then applies this circuit to the input. This equivalence allows us to apply Yao's minimax principal:

\begin{lemma}[Application of Yao's Principal]\label{lem:yao}
Assume there exists an SNN $\mathcal{N}$ with $\ell$ auxiliary neurons, which given any inputs $X \in \{0,1\}^n$ and $Y \in \{0,1\}^{\log n}$ which are fixed for rounds $\{0,...,t\}$, solves indexing with probability $\ge 1- \delta$ in $t$ rounds. Then there exists a feedforward deterministic linear threshold circuit $\mathcal{N}_D$ with $(t-1)\cdot (\ell + 1)$ auxiliary gates which solves indexing with probability $\ge 1-\delta$ given $X,Y$ drawn uniformly at random.
\end{lemma}
\begin{proof}
This follows from Yao's principal \cite{yao1977probabilistic}. In short, given $X,Y$ drawn uniformly at random, $\mathcal{N}$ solves indexing with probability $\ge 1-\delta$ (since by assumption, it succeeds with this probability for any $X,Y$). By Lemma \ref{lem:det}, $\mathcal{N}$ performs identically to an algorithm which selects a deterministic circuit from some distribution $\mathcal{D}$ and then applies it to the input. So at least one circuit in the support of $\mathcal{D}$ must succeed with probability $\ge 1-\delta$ on $X,Y$ drawn uniformly at random, since the success probability of $\mathcal{N}$ on the uniform distribution is just an average over the deterministic success probabilities.
\end{proof}

From Lemma \ref{lem:yao} we have a corollary which concludes our reduction from our spiking sigmoid lower bound to a lower bound on deterministic indexing circuits.
\begin{corollary}[Reduction to Deterministic Indexing on a Subset of Inputs]\label{cor:finalReduction}
Assume there exists an SNN $\mathcal{N}$ with $\ell$ auxiliary neurons, which, given inputs $X \in \{0,1\}^n$ and $Y \in \{0,1\}^{\log n}$ which are fixed for rounds $\{0,...,t\}$, solves indexing with probability $\ge 1- \frac{1}{2n}$ in $t$ rounds. Then there exists some subset of inputs $\mathcal{X} \subseteq \{0,1\}^n$ with $|\mathcal{X}| \ge 2^{n-1}$ and a feedforward deterministic linear threshold circuit $\mathcal{N}_D$ with $(t-1) \cdot (\ell + 1)$ auxiliary gates which solves indexing given any $X \in \mathcal{X}$ and any index $Y \in \{0,1\}^{\log n}$.
\end{corollary}
\begin{proof}
Applying Lemma \ref{lem:yao} yields $\mathcal{N}_D$ which solves indexing on uniformly random  $X,Y$ with probability $1-\frac{1}{2n}$. Let $\mathbb{I}(X,Y) = 1$ if $\mathcal{N}_D$ solves indexing correctly on $X,Y$ and $0$ otherwise. Then:
\begin{align*}
1 - \frac{1}{2n} &\le 
\frac{1}{n \cdot 2^n} \sum_{X \in \{0,1\}^n} \sum_{Y \in \{0,1\}^{\log n}} \mathbb{I}(X,Y)
= \E_{X \text{ uniform from }  \{0,1\}^n} \left [ \frac{1}{n}\sum_{Y \in \{0,1\}^{\log n}} \mathbb{I}(X,Y) \right ]
\end{align*}
which in turn implies:
\begin{align}\label{eq:expBound}
 \E_{X \text{ uniform from }  \{0,1\}^n} \left [ \frac{1}{n}\sum_{Y \in \{0,1\}^{\log n}} (1- \mathbb{I}(X,Y)) \right ] \le \frac{1}{2n}.
\end{align}

If $\frac{1}{n}\sum_{Y \in \{0,1\}^{\log n}} (1- \mathbb{I}(X,Y)) \neq 0$ then $\frac{1}{n}\sum_{Y \in \{0,1\}^{\log n}} (1- \mathbb{I}(X,Y))  \ge \frac{1}{n}$ just by the fact that the sum is an integer. Thus, for \eqref{eq:expBound} to hold, we must have $\frac{1}{n}\sum_{Y \in \{0,1\}^{\log n}} (1- \mathbb{I}(X,Y)) = 0$ for at least $\frac{1}{2}$ of the inputs $X \in \{0,1\}^n$. That is, $\mathcal{N}_D$ solves indexing for every input index on some subset $\mathcal{X}$ with $|\mathcal{X}| \ge \frac{1}{2} | \{0,1\}^n| \ge 2^{n-1}$.
\end{proof}

\subsection{Lower Bound for Deterministic Indexing on a Subset of Inputs}

With Corollary \ref{cor:finalReduction} in place, we now turn to lower bounding the size of a deterministic linear threshold circuit $\mathcal{N}_D$ which solves the indexing problem on some subset of inputs $\mathcal{X}$ with $|\mathcal{X}| \ge 2^{n-1}$. To do this, we employ VC dimension techniques first introduced for bounding the size of linear threshold circuits computing indexing on all inputs \cite{koiran1996vc}.

Consider fixing some input $X \in \mathcal{X}$, such that the output of $\mathcal{N}_D$ is just a function of the index $Y$. Specifically, with $X$ fixed, $\mathcal{N}_D$ computes the function $f_X : \{0,1\}^{\log n} \rightarrow \{0,1\}$ whose truth table is given by $X$. Note that the output of $\mathcal{N}_D$ with $X$ fixed is equivalent to the output of a feedforward linear threshold circuit $\mathcal{N}_D^X$ where each gate with an incoming edge from $x_i \in X$ has its threshold adjusting to reflect the weight of this edge if $x_i = 1$. 

We define  two sets of functions. Let $\mathcal{F} = \{f_X | X \in \mathcal{X} \}$ be all functions computable using some $\mathcal{N}_D^X$ as defined above. Further, let $\mathcal{G}$ be the set of all functions computabled by any circuit $\mathcal{N}_D'$ which is generated by removing the input gates of $\mathcal{N}_D$ and adjusting the threshold on each remaining gate to reflect the effects of any inputs with $x_i = 1$. We have $\mathcal{F} \subseteq \mathcal{G}$ and hence, letting $VC(\cdot)$ denote the VC dimension of a set of functions have: 
$VC(\mathcal{F}) \le VC(\mathcal{G}).$
We can now apply two results. The first gives a lower bound $VC(\mathcal{F})$:
\begin{lemma}[Corollary 3.8 of \cite{anthony2009neural} -- Consequence of Sauer's Lemma \cite{sauer1972density,shelah1972combinatorial}]\label{lem:sauer}
For any set of boolean functions $\mathcal{H} = \{h \}$ with $h: \{0,1\}^{\log n} \rightarrow \{0,1\}$:
$$VC(\mathcal{H}) \ge \frac{\log |\mathcal{H}|}{\log n + \log e}.$$
\end{lemma}
We next upper bound $VC(\mathcal{G})$. We have the following, whose proof is in Appendix \ref{sec:mislb}:

\begin{lemma}[Linear Threshold Circuit VC Bound]\label{lem:circuitVC}
Let $\mathcal{H}$ be the set of all functions computed by a fixed feedforward linear threshold circuit with $m \ge 2$ gates (i.e. fixed edges and weights), where each gate has a variable threshold. Then:
$VC(\mathcal{H}) \le 3 m \log m.$
\end{lemma}
\def\APPENDLTCVC{
We first define a generalization of VC dimension, which measures the number of dichotomies that a class of functions can induce over a set.
\begin{definition}\label{def:dich}
For a class of functions $\mathcal{H}: X \rightarrow \{0,1\}$, let $\Delta_{\mathcal{H}}(z)$ be the maximum over sets $A \subseteq X$ with $|A| = z$ of $\Delta_\mathcal{H}(A)$, the number of different partitions of $A$ that can be induced by some $h \in \mathcal{H}$. $VC(\mathcal{H})$ is the maximum $z$ with $\Delta_{\mathcal{H}}(z) = 2^z$.
\end{definition}

\begin{lemma}[Theorem 1 of \cite{baum1989size}]\label{thm1}
Let $\mathcal{H}$ be the class of functions computed by a fixed feedforward architecture with $m$ nodes, where each node $v_i$ can be chosen to compute any function in the class $\mathcal{H}_i$.
Then $\Delta_\mathcal{H}(z) \le \prod_{i=1}^m \Delta_{\mathcal{H}_i}(z)$.
\end{lemma}
Applying Lemma \ref{thm1} to a fixed threshold circuit with programmable thresholds gives Lemma \ref{lem:circuitVC}.
For a threshold gate with fixed input weights and a variable threshold, we trivially have
 $\Delta_{\mathcal{H}_i}(z) \le z$, since for $z$ inputs, there are $z$ possible functions that can computed by varying the threshold of the circuit (since edge weights are fixed, any input is just mapped to a single real number which is thresholded). We thus have $\Delta_{\mathcal{H}}(z) \le z^m$.
 
 By Definition \ref{def:dich}, for $z = VC(\mathcal{H})$ we have $\Delta_{\mathcal{H}}(z) = 2^{z}.$ So we have $2^z \le z^m$ and thus $z \le m \log z$.
This is violated for any $z \ge 3m \log m$, so we must have $VC(\mathcal{H}) \le 3 m \log m$.
}
Applying the bounds of Lemmas \ref{lem:sauer} and \ref{lem:circuitVC} along with $VC(\mathcal{F}) \le VC(\mathcal{G})$ gives:
\begin{lemma}[Deterministic Circuit Lower Bound]\label{lem:detLB}
For any set $\mathcal{X} \subseteq \{0,1\}^n$ with $|\mathcal{X}| \ge 2^{n-1}$, any feedforward deterministic linear threshold circuit $\mathcal{N}_D$ with $m$ non-input gates which solves indexing given any $X \in \mathcal{X}$ and any index $Y \in \{0,1\}^{\log n}$ must have
$m = \Omega \left ( \frac{n}{\log^2 n} \right ).$
\end{lemma}
\begin{proof}
Let $\mathcal{F}$ and $\mathcal{G}$ be as defined in the beginning of the section. We have $VC(\mathcal{F}) \le VC(\mathcal{G})$. At the same time, by Lemma \ref{lem:sauer} we have
$$VC(\mathcal{F}) \ge \frac{\log |\mathcal{F}|}{\log n + \log e} = \frac{\log |\mathcal{X}|}{\log n + \log e} \ge \frac{c n}{\log n}$$
for some fixed constant $c$. By Lemma \ref{lem:circuitVC} we have
$VC(\mathcal{G}) \le 3 m \log m.$
We thus can conclude that $\frac{c n}{\log n} \le 3 m \log m$, and so $m = \Omega \left (\frac{n}{\log^2 n} \right).$
\end{proof}
We conclude by proving our main lower bound:
\begin{proof}[Proof of Theorem \ref{thm:lower}]
The existence of a spiking sigmoidal network with $\ell$ auxiliary neurons, solving indexing with probability $\ge 1- \frac{1}{2n}$ in $t$ rounds implies via Corollary \ref{cor:finalReduction} the existence of a feedforward deterministic linear threshold circuit with $(t-1)\ell + 1$ non-input gates solving indexing on some subset of inputs $\mathcal{X}$ with $|\mathcal{X}| \ge 2^{n-1}$. Thus by Lemma \ref{lem:detLB} we must have $\ell \cdot t = \Omega \left (\frac{n}{\log^2 n} \right )$.
\end{proof}

\vspace{-2em}
\section{Applications to Similarity Testing and Compression}\label{sec:applications}
\subsection{Similarity Testing}\label{sec:equality}

\begin{theorem}[Similarity Testing]\label{thm:equality} There exists an SNN with $O \left (\frac{\sqrt{n} \log n}{\epsilon}\right)$ auxilary neurons that solves the approximate equality testing problem in $O (\sqrt{n})$ rounds. Specifically, given inputs $X_1,X_2 \in \{0,1\}^n$  which are fixed for all rounds $t \in \{0,..,5\sqrt{n}+2\}$, the output $z$ satisfies \whp $z^{5\sqrt{n}+2} = 1$ if $\ham(X_1, X_2) \ge \epsilon n$. Further if $X_1 = X_2$ then $z^{5\sqrt{n}+2}  = 0$ \whp
\end{theorem}

\setlength{\columnsep}{17pt}%
\begin{wrapfigure}{r}{7cm}
\includegraphics[width=0.38\textwidth]{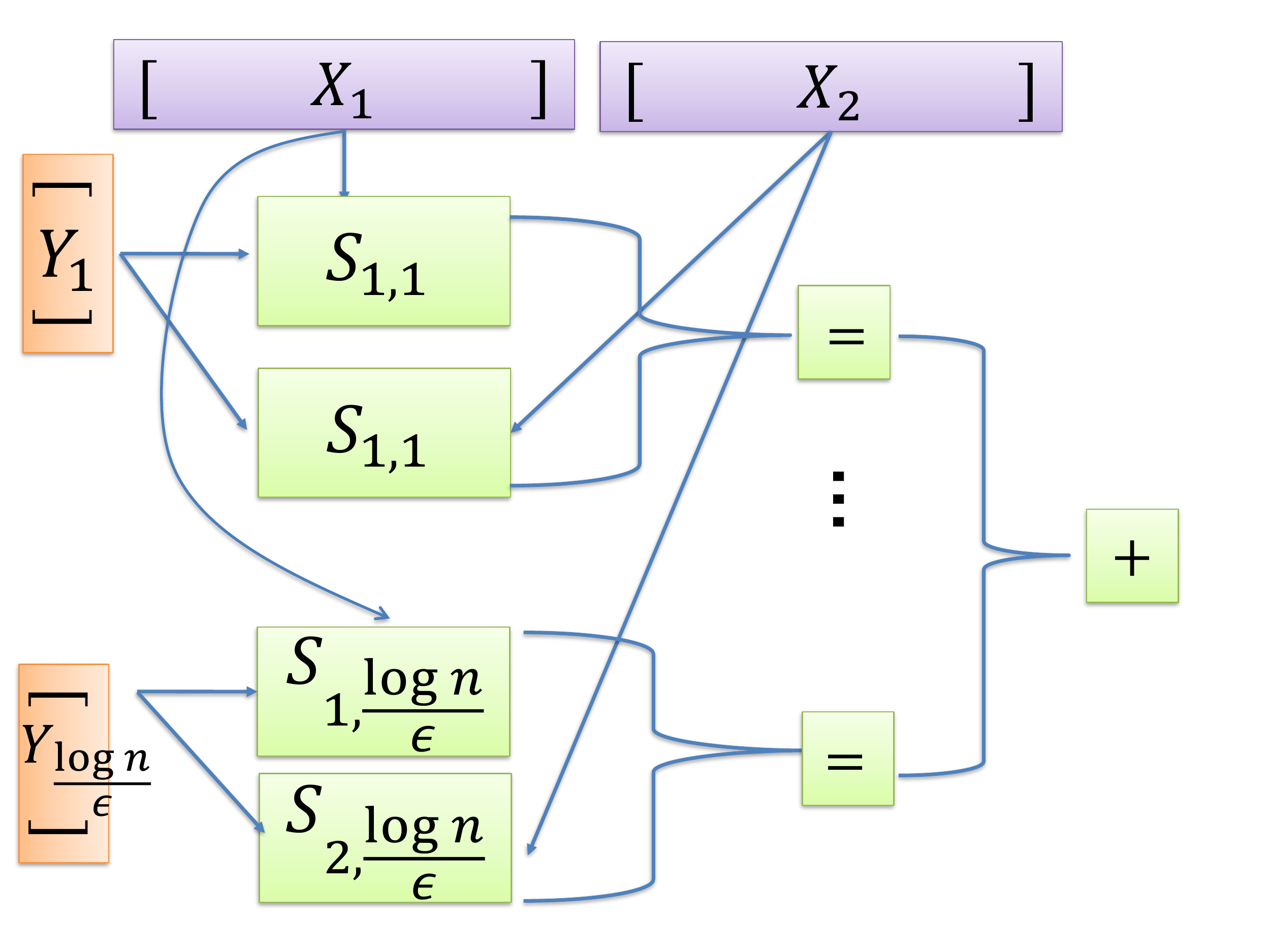}
\vspace{-1em}
	\caption{\footnotesize Solving $\epsilon$-approximate similarity using Neuro-RAM modules.}
	\vspace{-2em}
	\label{fig:equality}
\end{wrapfigure}

Our similarity testing network will use $K = \Theta \left (\frac{\log n}{\epsilon} \right)$ copies of our Neuro-RAM from Theorem \ref{thm:selection}, labeled $S_{1,k}$ and $S_{2,k}$ for all $k \in \{1,...,K\}$. 
The idea will be to employ $\log n$ auxiliary neurons $Y_k = y_{1,k},...,y_{\log n,k}$ whose values encode a \emph{random index} $i \in \{0,...,n-1\}$. 
By feeding the inputs $(X_1,Y_k)$ and $(X_2,Y_k)$ into $S_1$ and $S_2$, we can check whether $X_1$ and $X_2$ match at position $i$. Checking $\Theta \left (\frac{\log n}{\epsilon} \right )$ different random indices suffices that identify if $\ham(X_1,X_0) \ge \epsilon n$ \whp Additionally, if $X_1 = X_0$, they will never differ at any of the checks, and so the output will never be triggered. We use the following:

\begin{observation}\label{obs:2test} Consider $X_1,X_2 \in \{0,1\}^n$ with $\ham(X_1,X_0) \ge \epsilon n$. Let $i_1,...,i_T$ be chosen independently and uniformly at random in $\{0,...,n-1\}$. Then for $T = \frac{c \ln n}{\epsilon}$,
$$\Pr [ (X_1)_{i_t} = (X_2)_{i_t}\text{ for all }t \in 1,...,T] \le \frac{1}{n^c}.$$
\end{observation}
\begin{proof}
For any fixed $t$, $\Pr [ (X_1)_{i_t} = (X_2)_{i_t}] = 1- \frac{\epsilon n}{n} = 1- \epsilon$ as we select indices at random. Additionally, each of these events is independent since $i_1,...i_T$ are chosen independently so:
$
\Pr [ (X_1)_{i_t} = (X_2)_{i_t}\text{ for all }t \in 1,...,T] \le (1-\epsilon)^T =\left (1-\epsilon)^{1/\epsilon}\right )^{c \ln n} \le \frac{1}{e^{c \ln n}} \le \frac{1}{n^c}.
$
\end{proof}

\subsubsection{Implementation Sketch}

It is clear that the above strategy can be implemented in the spiking sigmoidal network model -- we sketch the construction here. By Theorem \ref{thm:selection}, we require $O \left (\frac{\sqrt{n}\log n}{\epsilon} \right )$ auxiliary neurons for the $2K = \Theta \left (\frac{\log n}{\epsilon} \right)$ neuro-RAMs employed, which dominates all other costs.

It suffices to present a random index to each pair of neuro-RAMs $S_1,k$ an $S_2,k$ for $5\sqrt{n}$ rounds (the number of rounds required for the network of Theorem \ref{thm:selection} to process an $n$-bit input). To implement this strategy, we need two simple mechanisms, described below.

\paragraph{Random Index Generation:}
For each of the $\log n$ index neurons in $Y_k$ we set $\Bias(y_i) = 0$ and add a self-loop $w(y_i,y_i) = 2$. In round $1$, since they have no-inputs, each neuron has potential $0$ and fires with probability $1/2$. Thus, $Y_k^1$ represents a random index in $\{0,...,n-1\}$. To propagate this index we can use a single auxiliary inhibitory neuron $g$, which has bias $\Bias(g) = 1$ and $w(x,g) = 2$ for every input neuron $x$. Thus, $g$  fires \whp in round $1$ and continues firing in all later rounds, as long as at least one input fires.

We add an inhibitory edge from $g$ to $y_i$ for all $i$ with weight $w(g,y_i) = -1$.
 The inhibitory edges from $g$ will keep the random index `locked' in place. The inhibitory weight of $-1$ prevents any $y_i$ without an active self-loop from firing \whp but allows any $y_i$ with an active self-loop to fire \whp since it will still have potential $\Bias(y_i) + w(y_i,y_i) - 1 = 1$.
 
 If both inputs are $0$, $g$ will not fire \whp However, here our network can just output $0$ since $X_1 = X_2$ so it does not matter if the random indices stay fixed.
 
 

\paragraph{Comparing Outputs:}
We next handle comparing the outputs of $S_{1,k}$ and $S_{2,k}$ to perform equality checking. 
We use two auxiliary neurons -- $f_{1,k}$ and $f_{2,k}$. $f_{1,k}$ is excitatory and fires \whp as long as long as at least one of $S_{1,k}$ or $S_{2,k}$ has an active output. $f_{2,k}$ is an inhibitor that fires only if \emph{both} $S_{1,k}$ and $S_{2,k}$ have active outputs. We then connect $f_{1,k}$ to our output $z$ with weight $w(f_{1,k},z) = 2$ and connect $f_{2,k}$ with weight $w(f_2,z) = -2$ for all $k$. We set $\Bias(z) = 1$. In this way, $z$ fires in round $5\sqrt{n}+2$ \whp if for some $k$, \emph{exactly one} of $S_{1,k}$ or $S_{2,k}$ has an active output in round $5\sqrt{n}$ and hence an inequality is detected. Otherwise, $z$ does not fire \whp This behavior gives the output condition of Theorem \ref{thm:equality}.

\subsection{Randomized Compression}

We conclude by discussing informally how our neuro-RAM can be applied beyond similarity testing to other randomized compression schemes. Consider the setting where we are given $n$ input vectors $X_i \in \{0,1\}^d$. Let $\bv{X} \in \{0,1\}^{n \times d}$ denote the matrix of all inputs. Think of $d$ as being a large ambient dimension, which we would like to reduce before further processing. 

One popular technique is \emph{Johnson-Lindenstrauss (JL) random projection}, where $\bv{X}$ is multiplied by a random matrix $\bs{\Pi} \in \mathbb{R}^{d \times d'}$ with $d' << d$ to give the compressed dataset $\bv{\tilde X} = \bv{X} \bs{\Pi}$. \emph{Regardless of the initial dimension $d$}, if $d'$ is set large enough, $\bv{\tilde X}$ preserves significant information about $\bv{X}$. $d' = \tilde O(\log n)$ is enough to preserve the distances between all points \whp \cite{kane2014sparser}, $d' = \tilde O(k)$ is enough to use $\bv{\tilde X}$ for approximate $k$-means clustering or $k$-rank approximation \cite{boutsidis2010random,cohen2015dimensionality}, and $d' = \tilde O(n)$ preserves the full covariance matrix of the input and so $\bv{\tilde X}$ can be used for approximate regression and many other problems \cite{clarkson2013low,sarlos2006improved}.

JL projection has been suggested as a method for neural dimensionality reduction \cite{allen2014sparse,ganguli2012compressed}, where $\bs{\Pi}$ is viewed as a matrix of random synapse weights, which connect the input neurons representing $\bv{X}$ to the output neurons representing $\bv{\tilde X}$. While this view is quite natural, we often want to draw $\bs{\Pi}$ with \emph{fresh randomness} for each input $\bv{X}$. This is not possible using changing synapse weights, which evolve over a relatively long time scale. Fortunately, it is possible to simulate these random connections using our neuro-RAM module.

Typically, $\bs{\Pi}$ is sparse so that it can be multiplied by efficiently.
 In one of the most efficient constructions \cite{clarkson2013low}, it has just a single nonzero entry in each row which is chosen randomly to be $\pm 1$ and placed in a uniform random position in the row.  Thus, computing a single bit of $\bv{\tilde X} = \bv{X}\bs{\Pi}$ requires selecting on average $d/d'$ random columns of $\bv{X}$, multiplying their entries by a random sign and summing them together. This can be done with a set of neuro-RAMS, each using $O(\sqrt{d})$ auxiliary neurons which select the random columns of $\bv{X}$. In total we will needs $\tilde O(d/d')$ networks -- the maximum column sparsity of $\bs{\Pi}$ with high probability, yielding $O(d^{3/2}/d')$ auxiliary neurons total. In contrast, a naive simulation of random edges using spiking neurons would require  $\Theta(d)$ auxiliary neurons, which is less efficient whenever $d' > d^{3/2}$. Additionally, our neuro-RAMs can be reused to compute multiple entries of $\bv{\tilde X}$, which is not the case for the naive simulation.
 
Traditionally, the value of an entry of $\bv{\tilde X}$ is a real number, which cannot be directly represented in a spiking neural network. In our construction, the value of the entry is encoded in its potential, and we leave as an interesting open question how this potential should be decoded or otherwise used in downstream applications of the compression.

\subsection*{Acknowledgments} We are grateful to Mohsen Ghaffari. Some of the ideas of this paper came up while visiting him at ETH. We would like to thank Sergio Rajsbaum, Ron Rothblum and Nir Shavit for helpful discussions.

\bibliography{selection}{}
\bibliographystyle{alpha}

\newpage
\appendix

\section{Missing Details for Implementing Neuro-RAM}
\label{sec:neuroram}
\APENDSELECTION
\APPENDMANYA

\section{Missing Proofs for the Lower Bound}\label{sec:mislb}
\textbf{Proof of Lemma \ref{lem:circuitVC}:}
\APPENDLTCVC

\end{document}